\documentclass[a4paper,12pt]{article}
\usepackage[francais]{babel}
\usepackage{inputenc} 
\usepackage[T1]{fontenc}       
\usepackage{graphicx}
\usepackage{lfa08}  
\usepackage{times}
\usepackage{amssymb, amsthm, amsmath, amsfonts}

\newcommand{\ind}{\mbox{1\hspace{-.25em}l}}

\newcommand{\conj}{\mathrm{Conj}}

\newcommand{\DS}{\mathrm{DS}}

\newcommand{\bel}{\mathrm{bel}}
\newcommand{\pl}{\mathrm{pl}}
\newcommand{\betP}{\mathrm{betP}}
\newcommand{\Bel}{\mathrm{Bel}}
\newcommand{\Pl}{\mathrm{Pl}}
\newcommand{\GPT}{\mathrm{GPT}}

\newcommand{\argmax}{\operatornamewithlimits{arg\,max}}

\begin{document}

\renewcommand{\figurename}{Figure~}
\renewcommand{\tablename}{Tableau~}
\date{}

\title{ Aide à la décision crédibiliste et rejet pour la reconnaissance d'images texturées \\[1.3em]
        Belief decision support and reject for textured images characterization
}

\author{\small
        A. Martin \footnote{Ce travail a été réalisé lors d'un séjour au RDDC (Recherche et Développement pour la Défence Canada) à Valcatier, Québec, Canada, et est en partie financé par la DGA (Délégation générale pour l'Armement) et BMO (Brest Métropole Océane).} \\
        ~\\
            \small{ENSIETA, E$^3$I$^2$ EA3876} 
        ~\\
        \\
        \small{2 rue François Verny 29806 Brest Cedex 9, Arnaud.Martin@ensieta.fr}
}

\parskip 3mm
\maketitle

\thispagestyle{empty}

\doubleresume{
La classification d'images texturées suppose considérer les images en région de même texture pour les classifier. En environnement incertain, il peut être préférable de prendre une décision imprécise ou de rejeter la région correspondant à une classe non apprise. De plus, sur les unités de classification que sont les régions, peut apparaître plus d'une texture. Ces considérations nous ont entraîné à développer un modèle de décision crédibiliste permettant de rejeter une région comme non apprise et de décider sur les unions et intersections de classes apprises. 

L'approche proposée trouve toute sa justification dans une application de caractérisation de fond marin à partir d'images sonar, qui sert d'illustration.}
{Décision crédibiliste, rejet, théorie des fonctions de croyance, reconnaissance, images texturées.}
{The textured images' classification assumes to consider the images in terms of area with the same texture. In uncertain environment, it could be better to take an imprecise decision or to reject the area corresponding to an unlearning class. Moreover, on the areas that are the classification units, we can have more than one texture. These considerations allows us to develop a belief decision model permitting to reject an area as unlearning and to decide on unions and intersections of learning classes.

The proposed approach finds all its justification in an application of seabed characterization from sonar images, which contributes to an illustration.}
{Belief decision, reject, belief function theory, pattern recognition, textured images.}

\section{Introduction}
La classification d'images texturées en environnements incertains pose plusieurs problèmes. Tout d'abord, caractériser la texture d'une image nécessite de considérer une région supposée contenir une même texture sur laquelle différents paramètres de texture pourront être calculés selon de nombreuses méthodes. Généralement le découpage de l'image se fait en imagettes de taille carrée (par exemple $32 \times 32$ pixels) qui dépend de la texture à caractériser et de la résolution de l'image. Cette taille est donc fixée selon ces caractéristiques de l'application. Le découpage de l'image est réalisé en déplaçant l'imagette éventuellement avec un pas de recouvrement sur toute l'image. Une imagette peut donc contenir deux ou plus textures différentes. Chercher à caractériser des zones (de forme variée) de texture homogène nécessite de déterminer les contours de ces régions ce qui peut s'avérer très difficile en environnement incertain lorsque ces contours sont flous \cite{Martin06}. 

Ainsi, l'unité de classification est l'imagette. Pour l'apprentissage du classifieur, il est donc nécessaire de considérer des imagettes contenant une seule texture. Chercher à apprendre sur des imagettes contenant plusieurs textures (et donc considérer davantage de classes que de texture) pose problème~: le nombre de textures sur une imagette peut être important et la  proportion de chaque texture présente sur l'imagette varie de manière continue. 

Nous avons montré l'intérêt de la théorie des fonctions de croyance pour la prise de décision lors de la classification d'images texturées en environnement incertain \cite{Martin08}. L'application visée et qui servira d'illustration dans ce papier est la caractérisation de sédiments marins à partir d'images sonar. Ce problème est particulièrement délicat car les images sonar sont souvent difficiles à interpréter, même pour un expert \cite{Martin06}. Sur une même imagette, il peut être préférable de décider qu'il y a par exemple du sable ou de la vase lorsque l'expert est peu sûr ou de rejeter si l'imagette correspond à une classe non apprise (telle qu'une épave) \cite{Martin08}. Il faudrait également pouvoir décider qu'il y a du sable et de la vase lorsque l'expert est sûr et que sur l'imagette considérée les deux sédiments apparaissent. Nous supposons donc posséder un classifieur s'exprimant par une fonction de croyance. Nous considérerons ici la combinaison de classifieurs binaires de type SVM (Support Vector Machine) présentée dans \cite{Martin08}, d'autres approches de combinaison de classifieurs sont possibles \cite{Martin05,Quost07, Aregui07} ou de classifieurs crédibilistes \cite{Denoeux95,Laanaya06}.

Nous présentons dans la section suivante un rappel sur les fonctions de croyance et le processus de décision utilisé dans \cite{Martin08} permettant de décider sur les classes et les unions de classes mais aussi de rejeter des éléments. Nous étendons ensuite cette approche afin de pouvoir décider de façon simple également sur les intersections de classes. Ceci est illustré dans la section \ref{illustration} pour décider à l'issue de la reconnaissance de la texture des images sonar.

\section{Fonctions de croyance}

La théorie des fonctions de croyance est de plus en plus employée pour la modélisation des incertitudes et imprécisions. Elle est fondée sur la manipulation des fonctions de masse (ou masse élémentaire de croyance). Les fonctions de masse sont définies sur l'ensemble de toutes les disjonctions du cadre de discernement $\Theta=\{C_1,\ldots,C_n\}$ et à valeurs dans $[0,1]$, où les $C_i$ représentent les hypothèses supposées exhaustives et exclusives. Cet ensemble est noté $2^\Theta$. Généralement, il est ajouté une condition de normalité, donnée par :
\begin{eqnarray}
\label{hyp1_fonction_masse}
\sum_{X \in 2^\Theta} m_j(X)=1,
\end{eqnarray}
où $m_j(.)$ représente la fonction de masse pour une source (ou un classifieur binaire dans cet article) $S_j$, \linebreak $j=1,...,s$. Ainsi selon les stratégies du passage à plusieurs classes du classifieur binaire nous avons $s=n$ dans le cas un-contre-reste et $s=n(n-1)/2$ dans le cas un-contre-un.

\`A partir de ces fonctions de masse, d'autres fonctions de croyance peuvent être définies, telles que les fonctions de crédibilité, représentant une croyance minimale d'une source en un élément. Elles sont données pour tout $X \in 2^\Theta$ par~:
\begin{eqnarray}
\bel_j(X)=\sum_{Y \in 2^\Theta, \,Y \subseteq X, Y \neq \emptyset} m_j(Y),
\end{eqnarray}
ou encore les fonctions de plausibilité, représentant une croyance maximale d'une source en un élément, données pour tout $X \in 2^\Theta$ par~:
\begin{eqnarray}
\pl_j(X)=\sum_{Y \in 2^\Theta, \,Y \cap X \neq \emptyset} m_j(Y).
\end{eqnarray}

Afin de conserver un maximum d'informations, il est préférable de rester à un niveau crédal ({\em i.e.} de manipuler des fonctions de croyance) pendant l'étape de combinaison des informations pour prendre la décision sur les fonctions de décision issues de la combinaison. La règle initialement proposée par Dempster, est une règle conjonctive normalisée définie pour $s$ classifieurs, pour tout $X \in 2^\Theta \smallsetminus \{\emptyset\}$ par~:
\begin{eqnarray}
\label{DS}
m_\DS(X)=\frac{m_\conj(X)}{1-m_\conj(\emptyset)}, 
\end{eqnarray}
avec
\begin{eqnarray}
\label{conj}
m_\conj(X)=\sum_{Y_1 \cap ... \cap Y_s = X} \prod_{j=1}^s m_j(Y_j),
\end{eqnarray}
où $Y_j \in 2^\Theta$ est la réponse du classifieur $j$, et $m_j(Y_j)$ la fonction de masse associée. $m_\conj(\emptyset)$ est généralement interprétée comme une mesure de conflit ou plus exactement comme l'inconsistance de la fusion. Cette règle de combinaison appliquée à des fonctions de masse pour lesquelles les seules éléments focaux sont les singletons (\emph{i.e.} des probabilités) est équivalente à l'approche bayésienne. Un état de l'art des nombreuses règles de combinaison issues de la théorie des fonctions de croyance, ainsi que de nouvelles règles prometteuses ont été proposés par \cite{Martin07}.

Si la décision prise par le maximum de crédibilité peut être trop pessimiste, la décision issue du maximum de plausibilité est bien souvent trop optimiste. Le maximum de la probabilité pignistique, introduite par \cite{Smets90b}, reste le compromis le plus employé. La probabilité pignistique est donnée pour tout $X \in 2^\Theta$, avec $X \neq \emptyset$ par~:
\begin{eqnarray}
\label{pignistic}
\betP(X)=\sum_{Y \in 2^\Theta, \, Y \neq \emptyset} \frac{|X \cap Y|}{|Y|} m_\DS(Y),
\end{eqnarray}
où $|X|$ représente le cardinal de $X$. En effet, nous avons $\bel(X)\leq \betP(X) \leq \pl(X)$ pour tout $X \in 2^\Theta \smallsetminus \{\emptyset\}$. De même qu'avec les approches bayésiennes, il n'existe pas en général de fonction optimale pour prendre la décision. 

Dans cette étude, nous souhaitons autoriser le rejet d'une partie des données comme n'appartenant pas à l'ensemble des classes apprises. De nombreuses approches en reconnaissance de forme sont envisageables afin de rejeter une observation appartenant à une classe non apprise (\emph{cf.} \cite{Frelicot04}). Dans le cadre des fonctions de croyance, une décision pessimiste est préférable. Le critère proposé par \cite{le_hegarat97}, consiste à choisir la classe $C_k$ pour une observation si~:
\begin{eqnarray}
\label{maxBelRejet}
\left\{
\begin{array}{l}
	\bel(C_k)=\displaystyle \max_{1\leq i \leq n} \bel(C_i),\\
	\bel(C_k) \geq \bel(C_k^c).\\
\end{array}
\right.
\end{eqnarray}
L'ajout de cette seconde condition par rapport au maximum de crédibilité, permet de ne prendre une décision que si celle-ci est non ambiguë, c'est-à-dire si nous croyons plus en la classe $C_k$ qu'en son contraire. 

En règle générale, la décision est prise sur les singletons du cadre de discernement \cite{Denoeux97,Smets05}. En effet,  les fonctions de crédibilité, plausibilité et probabilité pignistique sont des fonctions $f_d$ croissantes avec l'inclusion ($A\!\subset \! B$ implique $f_d(A)\leq f_d(B)$). Il est cependant possible de prendre une décision sur les unions de singletons en pondérant ces fonctions de décision par une fonction d'utilité dépendant de la cardinalité des éléments. Ainsi, l'approche proposée par \cite{Appriou05} considère les fonctions de plausibilité permettant de décider n'importe quel élément de $2^\Theta$ et non plus seulement les singletons comme précédemment. Ainsi nous allons choisir l'élément $A \in 2^\Theta$ pour une observation si~:
\begin{eqnarray}
\label{DecAppriou}
	A=\argmax_{X \in 2^\Theta} \left(m_d(X)\pl(X)\right),
\end{eqnarray}
où $m_d$ est une masse définie par~:
\begin{eqnarray}
m_d(X)=K_d \lambda_X \left(\frac{1}{|X|^r}\right),
\end{eqnarray}
$r$ est un paramètre appartenant à $[0,1]$ permettant de choisir une décision allant du choix d'un singleton ($r=1$) à l'indécision totale ($r=0$). $\lambda_X$ permet d'intégrer le manque de connaissance sur l'un des éléments $X$ de $2^\Theta$. Dans cette étude nous poserons $\lambda_X=1$. La constante $K_d$ est un facteur de normalisation qui garantit la condition de l'équation (\ref{hyp1_fonction_masse}). Cette approche est envisageable également avec les fonctions de crédibilité ou probabilité pignistique. 

Dans \cite{Martin08}, nous avons proposé une stratégie permettant à la fois prendre une décision sur une union de classes ({\em i.e.} lorsque l'indécision est grande et que nous ne pouvons décider entre deux classes particulières) et ne pas prendre de décision lorsque notre croyance en un singleton est trop faible. Cette règle de décision est donc composée de deux étapes~: 
\begin{enumerate}
\item la règle de décision du maximum de crédibilité avec rejet définie par l'équation (\ref{maxBelRejet}) est appliquée afin de déterminer les éléments n'appartenant pas aux classes apprises.
\item la règle de décision de l'équation (\ref{DecAppriou}) est ensuite appliquée aux éléments non rejetés.
\end{enumerate}
Une autre approche possible serait d'appliquer la règle de décision de l'équation (\ref{DecAppriou}), puis la règle de décision du maximum de crédibilité avec rejet sur les éléments imprécis classés sur les unions.

\section{Décision et intersections}

En vue de prendre une décision sur les intersections du cadre de discernement $\Theta$ de taille $n$, nous pourrions changer le cadre de discernement en un cadre de discernement élargi de taille $2^n-1$ correspondant à toutes les parties disjointes (hypothèse d'exclusivité) du diagramme de Venn. Le principal problème est alors combinatoire car l'ensemble des disjonctions de ce nouveau cadre de discernement est de taille $2^{2^n-1}$. De plus dans cet espace de nombreux éléments ne correspondent pas forcément à une réalité physique (par exemple l'élément $C_1 \smallsetminus \{C_2 \cup C_3 \}$ trouverait difficilement un sens dans notre application). Une autre approche proposée par \cite{Dezert02} consiste à fermer l'espace $\Theta$ par les opérateurs d'union et d'intersection, cet espace est noté $D^\Theta$.

\subsection{Simples extensions}
La cardinalité des éléments de $D^\Theta$ doit être redéfinie par le nombre de parties disjointes du diagramme de Venn et la cardinalité de $X$ est notée ${\cal C_M}(X)$ \cite{Smarandache04}. Les fonctions de crédibilité et plausibilité sont naturellement étendues par~:
\begin{eqnarray}
\label{bel}
\Bel(X)=\sum_{Y \in D^\Theta, \, Y\subseteq X} m(Y),
\end{eqnarray}
\begin{eqnarray}
\label{pl}
\Pl(X)=\sum_{Y \in D^\Theta, \, X \cap Y \neq \emptyset} m(Y). 
\end{eqnarray}
Il est à noter qu'alors la plausibilité n'a plus d'intérêt car dans $D^\Theta$ tous les éléments contiennent l'intersection de tous les éléments de $\Theta$. Cependant la plausibilité peut avoir un sens si on ajoute des contraintes sur des intersections possiblement vides. On ne se place pas dans ce cas dans cet article.

La probabilité pignistique pour $X \in D^\Theta$, avec $X \neq \emptyset$ s'écrit~:
\begin{eqnarray}
\label{betp}
\GPT(X)=\sum_{Y \in D^\Theta, Y \neq \emptyset} \frac{{\cal C_M}(X \cap Y)}{{\cal C_M}(Y)} m(Y).
\end{eqnarray}

Le critère proposé par \cite{le_hegarat97} peut difficilement s'étendre dans $D^\Theta$ puisque $D^\Theta$ n'est pas fermé par le passage au complémentaire. Il est toutefois possible de rejeter une partie des données considérées sur $2^\Theta$ (typiquement les singletons), puis de prendre une décision sur $D^\Theta$ pour les données restantes. L'approche proposée par \cite{Appriou05} étendue à $D^\Theta$ permet de prendre une décision sur n'importe quel élément de $D^\Theta$ tenant compte à la fois de la fonction de masse et de la cardinalité. Ainsi nous allons choisir l'élément $A \in D^\Theta$ pour une observation si~:
\begin{eqnarray}
\label{DecAppriouDSmT}
	A=\argmax_{X \in D^\Theta} \left(m_d(X)f_d(X)\right),
\end{eqnarray}
où $f_d$ est la fonction de décision retenue (crédibilité, plausibilité, probabilité pignistique, ...) et $m_d$ est la fonction de masse définie par~:
\begin{eqnarray}
\label{MasseBayes}
m_d(X)=K_d \lambda_X \left(\frac{1}{{\cal C_M}(X)^r}\right),
\end{eqnarray}
$r$ est toujours un paramètre dans $[0,1]$ permettant à présent une décision de l'intersection de tous les singletons ($r=1$) (à la place des singletons dans $2^\Theta$) jusqu'à l'indécision totale $\Theta$ ($r=0$). $\lambda_X$ et $K_d$ sont définis comme précédemment. Sans contrainte, il ne faut donc pas choisir la plausibilité pour $f_d$.

\subsection{Décision selon la spécificité}
\begin{figure}[htb]
  \begin{center}
     \includegraphics[height=6cm]{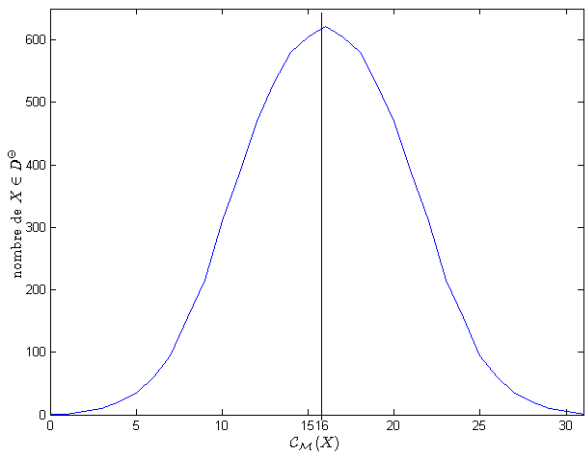}
  \end{center}
  \caption{Nombre d'éléments de $D^\Theta$ avec $|\Theta|=5$, pour une cardinalité donnée.}
  \label{distributionCardinalityDTheta5}
\end{figure}

La cardinalité ${\cal C_M}(X)$ peut être vue comme une mesure de spécificité de $X$. La figure~\ref{distributionCardinalityDTheta5} montre bien le rôle central joué par les singletons dans $D^\Theta$ (la cardinalité des singletons pour $|\Theta|$=5 est 16), mais aussi qu'il existe beaucoup d'autres éléments (619) ayant exactement la même cardinalité. Il peut être intéressant de préciser directement la spécificité (ou un intervalle de spécificité) des éléments sur lesquels nous souhaitons prendre une décision. C'est le rôle de $r$ dans l'approche de \cite{Appriou05}. Ainsi, nous décidons de l'élément $A$ si~:
\begin{eqnarray}
\label{extUnion}
	A=\argmax_{X \in {\cal S}} f_d(X),
\end{eqnarray}
où $f_d$ est la fonction de décision retenue (qui sera la fonction de décision pondérée par la masse donnée par l'équation~\eqref{MasseBayes} dans le cas d'un intervalle de spécificité) et
\begin{eqnarray}
{\cal S}\!\!=\!\!\left\{X\!\! \in \!\! D^\Theta ; min_S \! \leq \! {\cal C_M}(X)\! \leq \! max_S  \right\},
\end{eqnarray}
avec $min_S$ et $max_S$ respectivement le minimum et maximum de la spécificité attendue.

\section{Illustration}
\label{illustration}
\subsection{Images Sonar}
Notre base de données comporte 42 images sonar fournies par le GESMA (Groupe d'Etudes Sous-Marines de l'Atlantique). Ces images ont été obtenues à l'aide d'un sonar latéral Klein 5400 avec une résolution longitudinale de 20 à 30 cm et 3 cm en distance. Le fond se situe entre 15~m et 40~m.

Des experts ont segmenté manuellement ces images donnant le type de sédiment (roche, cailloutis, sable, vase, et ride (verticales ou à 45 degrés)), ainsi qu'ombre et autres qui ne seront pas pris comme classe. Il est cependant très difficile pour un expert de discriminer la roche des cailloutis et le sable de la vase. Pour le sédimentologue il est cependant important de pouvoir discriminer le sable de la vase. Le type ride peut être de sable ou de vase. Ainsi, avec le point de vue du sédimentologue, nous considérons seulement trois classes de sédiments~: $C_1$=roche-cailloutis, $C_2$=sable et $C_3$=vase. Dans le but d'évaluer notre processus de décision, nous considérons les rides comme une quatrième classe ($C_4$) qui ne sera pas apprise. 

Chaque image est découpée en imagettes de taille 32$\times$32 pixels (soit environ 40~m$^2$). L'apprentissage est réalisé à partir de 2333 imagettes homogènes de sable, vase et roche-cailloutis (soit 6999 imagettes homogènes). Le sédiment ride n'est pas appris volontairement. 

Afin de classifier ces imagettes, nous devons dans un premier temps extraire des paramètres de texture sur chacune d'elles. Après avoir tester différentes approches \cite{Martin05}, nous retenons ici les matrices de co-occurence qui sont calculées en comptabilisant le nombre de pixels de niveau de gris identique pour une distance et un angle fixés. Six paramètres proposés par Halarick sont alors calculés, représentant~: l'homogénéité, le contraste, l'entropie, la corrélation, la directivité et l'uniformité. La distance retenue est de 2 pixels et nous effectuons une moyenne sur les quatre directions~: 0, 45, 90 et 135 degrés. 

Nous utilisons la librairie {\it libSVM} \cite{Chang01} pour les classifieurs binaires avec une stratégie un-contre-un. Sur la base du taux de bonne classification, nous retenons le noyau gaussien (avec $\gamma=1/6$ où 6 représente la dimension de l'espace de nos données) avec une pondération de l'erreur $C=1$.

\subsection{Modèle}

Le modèle de fonctions de masse employé ici pour combiner les différents classifieurs binaires issus du SVM est celui proposé dans \cite{Martin08}. Dans le cas un-contre-un considéré dans cet article, les $n(n -1)/2$ fonctions de décision du classifieur sont notées $f_{ij}$ avec $i < j$ et $i,j=1,...,n$, où $i$ et $j$ correspondent aux classes considérées $C_i$ et $C_j$. Ainsi nous définissons la fonction de masse en sortie de chaque classifieur binaire par~:
\begin{eqnarray*}
\left\{
\begin{array}{l}
m_{ij} (C_i)(x)= \alpha_{ij}. \\
\quad  \left((1-\exp(-\frac{1}{\lambda_{ij,p}} f_{ij}(x)))\ind_{[0,+\infty[}(f_{ij}(x)) \right. \\
\quad \left. + \exp(-\frac{1}{\lambda_{ij,n}} f_{ij}(x)) \ind_{]-\infty,0[}(f_{ij}(x))\right) \\
\\
m_{ij} (C_j)(x)=  \alpha_{ij}.\\
\quad \left(\exp(-\frac{1}{\lambda_{ij,p}} f_{ij}(x))) \ind_{[0,+\infty[}(f_{ij}(x)) \right.\\
\quad \left. (1-\exp(-\frac{1}{\lambda_{ij,n}} f_{ij}(x))) \ind_{]-\infty,0[}(f_{ij}(x)) \right)\\
\\
m_{ij}(\Theta)(x)=  1-\alpha_{ij} \\
\end{array}
\right.
\end{eqnarray*}
avec
\begin{eqnarray}
\left\{
\begin{array}{l}
\lambda_{ij,p}=\displaystyle \frac{1}{l} \sum_{t=1}^{l} f_{ij}(x) \ind_{[0,+\infty[}(f_{ij}(x)),\\
\lambda_{ij,n}=\displaystyle \frac{1}{l} \sum_{t=1}^{l} f_{ij}(x) \ind_{]-\infty,0[}(f_{ij}(x)).
\end{array}
\right.
\end{eqnarray}

La règle de combinaison conjonctive normalisée (équation \eqref{DS}) sera ici employée sur les $n(n-1)/2$ fonctions de masse du cas un-contre-un. Lorsque les données sont très recouvertes, des règles répartissant le conflit plus finement doivent être privilégiés (\emph{cf.} \cite{Martin07}).

\subsection{Résultats et discussion}
Dans un premier temps donnons des résultats de reconnaissance sur des imagettes homogènes. On considère 1000 imagettes pour chaque type de sédiment roche-cailloutis ($C_1$), sable ($C_2$), vase ($C_3$), et ride ($C_4$). Les tableaux \ref{2TsansUnion} et \ref{2TUnion} donnent les résultats respectivement pour la décision pignistique et crédibilité avec rejet sur les singletons et décision selon \cite{Martin08} avec rejet puis sur les unions possibles. On constate tout d'abord l'intérêt du rejet, car même si une partie des imagettes des classes apprises est rejetée, la plus grande part rejetée vient des rides (classe $C_4$ non apprise). La part d'erreur la plus importante provient de la confusion entre sable ($C_2$) et vase ($C_3$) qui sont deux sédiments de texture homogène difficile à reconnaître pour l'expert \cite{Martin06}. Permettre de décider sur les unions montrent bien cette difficulté~: une part importante de ces imagettes de sable et vase sont classées comme sable ou vase ($C_2\cup C_3$). De plus le sédiment ride, s'il n'est pas rejeté, est classé en roche ou sable ($C_1\cup C_2$) qui correspond aux erreurs les plus commises. Les rides (de sable bien souvent) sont de texture plus ou moins hétérogène (comme la roche) et d'intensité proche du sable lorsque les rides sont peu marquées. 

\begin{table}[h]
\centering
  \begin{tabular}{|c|c|c|c||c|c|c|c|}
    \hline
    & \multicolumn{3}{c||}{pignistique} & \multicolumn{4}{|c|}{avec rejet} \\
    \hline
    & $\!\!C_1\!\!$ & $\!\!C_2\!\!$ & $\!\!C_3\!\!$ & $\!\!C_1\!\!$ &$\!\!C_2\!\!$ & $\!\!C_3\!\!$ & $\!\!C_4\!\!$\\
    \hline
    \!$C_1$\! & \!867 \! & \!131 \!& \! 2\! & \!825\! & \! 73\!  & \! 2\!  &\! 100\!\\
    \hline
   \!$C_2$\! & \!31\! & \!835\! & \!134\! & \!17 \! & \!655\!  & \! 98\!  & \!230\!\\
    \hline
    \!$C_3$\! & \!9\! & \!348\!  & \!643\! & \!4\! & \!216\! & \!590\! & \!190\!\\
    \hline
    \!$C_4$\! & \!567\! & \!415\! & \!18\! & \!486\! & \!232\! & \!13\! & \!269\!\\
    \hline
  \end{tabular}
\caption{Résultat de la décision pignistique et crédibilité avec rejet sur les singletons.}
\label{2TsansUnion}
\end{table}

Les résultats du tableau~\ref{2TUnion} sont donnés pour \linebreak $r=0.5$. Sur ces données homogènes, cette valeur de $r$ est un bon compromis entre singletons et unions \cite{Martin08}.
\begin{table*}[!htbp]
\centering
  \begin{tabular}{|c|c|c|c|c|c|c|c|c|}
    \hline
     & $C_1$ & $C_2$ & $C_3$ & $C_1 \cup C_2$ & $C_1 \cup C_3$ & $C_2 \cup C_3$ & $C_1 \cup C_2 \cup C_3$ & $C_4$\\
    \hline
    $C_1$  & 504 & 30 & 2 & 352 & 0 & 11 & 1 & 100\\
    \hline
   $C_2$  &  0 & 211 & 24 & 89 & 0 & 432 & 14 & 230\\
    \hline
   $C_3$  & 0 & 52 & 431 & 22 & 0 & 291 & 14 & 190 \\
    \hline
   $C_4$  & 131 & 104 & 6 & 420 & 0 & 49 & 21 & 269\\
    \hline
  \end{tabular}
\caption{Résultats avec rejet puis sur $2^\Theta$.}
\label{2TUnion}
\end{table*}

\`A présent si l'on prend la décision pignistique sur les éléments de cardinalité 4 dans $D^\Theta$, nous prenons donc la décision sur l'un des singletons ou $I_2=(C_1 \cap C_2)\cup (C_1\cap C_3) \cup (C_2\cap C_3)$. Le tableau~\ref{DTCm4} montre qu'on obtient exactement les mêmes résultats qu'avec la décision du maximum de crédibilité avec rejet (tableau~\ref{2TsansUnion}). L'hypothèse de monde fermé (et donc de rejet impossible) entraîne donc ici une interprétation d'une ignorance sur 2 types de sédiments sur une imagette ($I_2$).

\begin{table}[h]
\centering
  \begin{tabular}{|c|c|c|c|c|}
    \hline
    &  $\!\!C_1\!\!$ &$\!\!C_2\!\!$ & $\!\!C_3\!\!$ & $I_2$\\
    \hline
    $C_1$ &  \!825\! & \! 73\!  & \! 2\!  &\! 100\!\\
    \hline
   $C_2$ &  17  & 655  &  98  & 230\\
    \hline
    $C_3$ &  4 &  216 &  590 &  190\\
    \hline
    $C_4$ &  486  & 232  &  13 &  269\\
    \hline
  \end{tabular}
\caption{Résultat de la décision pignistique sur les éléments de cardinalité 4 dans $D^\Theta$.}
\label{DTCm4}
\end{table}

Considérons maintenant des imagettes ne contenant plus un seul sédiment, mais deux. Choisissons deux classes $S_1$ et $S_2$, correspondant à des types de sédiments appris ($S_1$=sable et roche-cailloutis et $S_2$=sable et vase). Prenons de plus deux classes $S_3$ et $S_4$ contenant des rides (non apprises), $S_3$=vase et ride et $S_4$=sable et ride. Nous testons chaque $S_i$ sur 299 imagettes hétérogènes. 

Si de nouveau nous comparons les résultats de la décision avec le maximum de crédibilité avec rejet et la décision pignistique sur les éléments de cardinalité 4, nous obtenons des résultats identiques donnés dans le tableau~\ref{DTCm4Het}.

\begin{table}[b]
\centering
  \begin{tabular}{|c|c|c|c|c|}
    \hline
    &  $\!\!C_1\!\!$ &$\!\!C_2\!\!$ & $\!\!C_3\!\!$ & $I_2$\\
    \hline
    $S_1$ &  44 & 166 & 10 & 79\\
    \hline
   $S_2$ &  0 & 55 & 145 & 99\\
    \hline
    $S_3$ &  71 & 71 & 54 & 103\\
    \hline
    $S_4$ &  69 & 150 & 3 & 77\\
    \hline
  \end{tabular}
\caption{Résultats sur les imagettes hétérogènes de la décision pignistique sur les éléments de cardinalité 4 dans $D^\Theta$.}
\label{DTCm4Het}
\end{table} 

Le tableau~\ref{DTCm2Het} montre les résultats avec et sans rejet pour la fonction de décision pignistique sur les éléments de cardinalité 2 (\emph{i.e.} ici les intersections de 2 classes). Rappelons que $S_1=C_1\cap C_2$ et $S_2=C_2 \cap C_3$. Nous n'obtenons aucun élément classé dans $C_1 \cap C_3$ ne correspondant à aucun $S_i$ (en trop faible nombre sur notre base). Ce résultat est plus dû à l'apprentissage des SVM et au choix de la fonction de masse que de l'approche de décision. Nous observons que $S_2$ est très bien classifiée en $C_2\cap C_3$, tandis que c'est moins clair pour $S_1$. Ceci peut s'expliquer par le fait que les paramètres de texture sont calculés sur l'imagette entière. Une imagette comprenant une partie de texture homogène et une autre hétérogène ne correspond à aucune texture apprise. Les imagettes de $S_4$ contenant des rides non apprises et de la vase, présentent une texture hétérogène et une luminosité moyenne proche du sable, ce qui peut expliquer les résultats. 

\begin{table}[h]
\centering
  \begin{tabular}{|c|c|c||c|c|c|}
    \hline
    & \multicolumn{2}{c||}{pignistique} & \multicolumn{3}{|c|}{avec rejet} \\
    \hline
    & $\!\!C_1\!\cap\!C_2\!\!$ & $\!\!C_2\!\cap\!C_3\!\!$ & $\!\!C_1\!\cap\!C_2\!\!$ & $\!\!C_2\!\cap\!C_3\!\!$ & $\!\!C_4\!\!$\\
    \hline
    $S_1$ & 162 & 137 & 103 & 117 & 79\!\\
    \hline
   $S_2$ &  45 & 254 & 26 & 174 & 99\\
    \hline
    $S_3$ & 172 & 127 & 101 & 95 & 103\\
    \hline
    $S_4$ & 202 & 97 & 130 & 92 & 77\\
    \hline
  \end{tabular}
\caption{Résultats sur les imagettes hétérogènes de la décision pignistique (avec et sans rejet) sur les éléments de cardinalité 2 dans $D^\Theta$.}
\label{DTCm2Het}
\end{table} 

Si l'on souhaite prendre une décision sur un intervalle de cardinalité, il est nécessaire de pondérer la fonction de décision par exemple avec la masse définie dans l'équation~\eqref{MasseBayes} en évitant de choisir la fonction de décision de plausibilité. 

Le tableau~\ref{DTCm2_6Het} donne les résultats obtenus sur les imagettes hétérogènes avec la décision du maximum de crédibilité pondérée par la masse de l'équation~\eqref{MasseBayes} sur les éléments de cardinalité comprise entre 2 et 6, ce qui correspond aux intersections de 2 classes jusqu'aux unions de 2 classes, avec $r=0.7$. On constate que la décision porte plus ici sur les singletons que sur les intersections, et aucune union n'est trouvée. Le type $S_1$ (roche et sable) est plus reconnue comme du sable, alors que $S_2$ (sable et vase) est reconnue pour moitié comme du sable et pour moitié comme de la vase. Les types $S_3$ (vase et ride) et $S_4$ (sable et ride) sont davantage reconnus comme du sable. 
\begin{table}[!h]
\centering
  \begin{tabular}{|c|c|c|c|c|c|}
    \hline
    & $C_1$ & $C_2$ & $C_3$ & $C_1 \cap C_2$ & $C_2 \cap C_3$ \\
    \hline
    $S_1$  & 55 & 227 & 12 & 3 & 2 \\
    \hline
   $S_2$  &  0 & 137 & 160 & 0 & 2\\
    \hline
   $S_3$  & 93 & 143 & 61 & 0 & 2 \\
    \hline
   $S_4$  & 88 & 207 & 3 & 1 & 0\\
    \hline
  \end{tabular}
\caption{Résultats sur les imagettes hétérogènes de la décision avec la crédibilité pondérée sur les éléments de cardinalité comprise entre 2 et 6 dans $D^\Theta$.}
\label{DTCm2_6Het}
\end{table}

\section{Conclusion}
La décision pour la reconnaissance d'images texturées en environnement incertain et particulièrement pour la reconnaissance de texture des fonds marins à partir d'images sonar est très délicate. Pour ce type d'applications, nous avons montré la possibilité et l'intérêt de pouvoir rejeter et décider sur les unions et intersections de classes apprises. L'approche proposée repose sur une extension d'approches existantes permettant d'une part de rejeter des données et d'autre part de décider sur des unions, étendues pour décider sur des intersections. Nous avons également fait ressortir l'intérêt de pouvoir spécifier la spécificité des éléments sur lesquelles on souhaite décider.

Ce type d'approche peut également avoir des intérêts pour découvrir des évènements rares, comme par exemple des épaves ou des mines non connues pour la chasse aux mines.

\end{document}